\documentclass[11pt]{article}
\usepackage{booktabs}

\usepackage[preprint]{acl}
\usepackage{amsmath}

\usepackage{times}
\usepackage{latexsym}
\usepackage{booktabs}
\usepackage{tabularx}
\usepackage{booktabs}
\usepackage{multirow}
\usepackage[table]{xcolor}
\usepackage{dblfloatfix}

\usepackage[T1]{fontenc}

\usepackage[utf8]{inputenc}

\usepackage{microtype}

\usepackage{inconsolata}

\usepackage{graphicx}
\usepackage{graphicx}
\usepackage{subcaption}

\title{Are Reasoning Vision-Language Models Robust to \\ Semantic Visual Distractions?}

\author{
 \textbf{Yizheng Sun\textsuperscript{1}},
 \textbf{Mochuan Zhan\textsuperscript{1}},
 \textbf{Yanan Ma\textsuperscript{1}},
 \textbf{Jia Tong See\textsuperscript{1}},
 \textbf{Yifan Wang\textsuperscript{1}},
 \\
 \textbf{Ziyi Wang\textsuperscript{2}},
 \textbf{Hao Li\textsuperscript{3}},
 \textbf{Yang Cui\textsuperscript{1}},
 \textbf{Wenhao Cai\textsuperscript{1}},
 \textbf{Jingyu Sun\textsuperscript{1}},
 \\
 \textbf{Chenghua Lin\textsuperscript{1}},
 \textbf{Riza Batista-Navarro\textsuperscript{1}},
 \textbf{Jingyuan Sun\textsuperscript{1,*}}
\\
\\
 \textsuperscript{1}University of Manchester,
 \textsuperscript{2}Marex,
 \textsuperscript{3}Imperial College London,
\\
 \small{
   \textbf{*Correspondence:} \href{jingyuan.sun@manchester.ac.uk}{jingyuan.sun@manchester.ac.uk}
 }
}

\begin{document}
\maketitle

\begin{abstract}
Reasoning Vision-Language Models (VLMs) achieve strong performance on complex multimodal tasks, but reliable real-world application requires handling visual inputs that are messier than clean, curated benchmarks. Existing works mainly evaluate such reliability of VLMs through input corruptions, such as noise, blur and weather effects, which make visual evidence harder to perceive. This leaves a critical reliability failure mode underexplored: a model may perceive the evidence correctly, yet reason from plausible but irrelevant and distracting evidence and propagate this mistake to its final answer. To address this gap, we introduce \textbf{Distract-Bench}, a benchmark for evaluating VLM robustness to \textbf{semantic visual distractions}, defined as meaningful but task-irrelevant visual cues added to inputs while preserving the ground-truth answer. We comprehensively evaluate eight leading open-source and two closed-source VLMs across conventional vision corruptions and Distract-Bench. Our results show that Distract-Bench exposes a robustness failure distinct from vision corruptions: reasoning VLMs largely track their non-reasoning base models under perceptual degradation, but show consistently lower robustness to semantic distractions. Further analysis shows that these distractions often enter the reasoning process of VLMs, are treated as evidence, and lead to incorrect answers. Together, these findings reframe robustness evaluation for reasoning VLMs, shifting the focus from degraded perception to distractions for reliable real-world visual reasoning. Our data and code are available at \href{https://github.com/Yizheng-Sun/Distract-Bench}{https://github.com/Yizheng-Sun/Distract-Bench}.
\end{abstract}

\section{Introduction}

\begin{figure*}[!h]
\centering
\includegraphics[width=\textwidth]{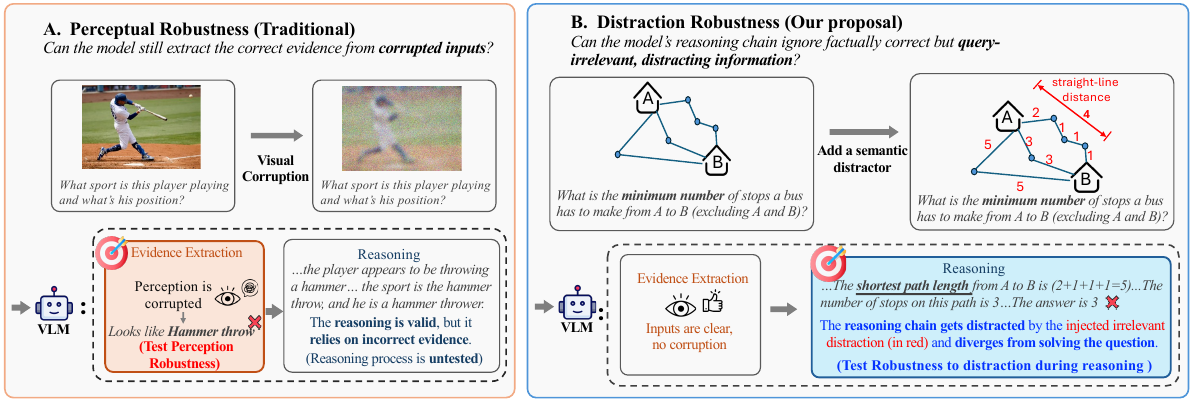}
\caption{\label{fig:intro_fig} \textbf{From perceptual robustness to semantic-distraction robustness.} Traditional robustness evaluation tests whether VLMs can still perceive evidence from visually corrupted inputs, where failures mainly reflect degraded perception. Distract-Bench targets a different failure mode: the image remains clear and the ground-truth answer is unchanged, but a factually correct, query-irrelevant semantic distractor is injected. This setting tests whether reasoning VLMs can select the right evidence and ignore plausible but irrelevant visual cues, or whether such distractors enter the reasoning chain and lead to incorrect answers.}
\end{figure*}

Large Vision-Language Models (VLMs) have advanced rapidly, achieving strong performance across a wide range of vision-language tasks \citep{liu2023visual,bai2025qwen3,hurst2024gpt}. More recently, inspired by reasoning-centric language models trained with reinforcement learning \citep{guo2025deepseek}, a growing line of work has equipped VLMs with explicit long-form reasoning abilities \citep{yang2025r1,huang2025vision,leng2025mmr1,meng2025mm}. These reasoning-enhanced VLMs have shown substantial gains on multimodal mathematical reasoning \citep{lu2024mathvista,wang2024measuring}, scientific understanding, and general visual question answering \citep{yue2024mmmu,mmstar}. This progress suggests that stronger reasoning is becoming a central path toward more capable multimodal systems.

Despite the strong performance of reasoning VLMs, reliable real-world application requires handling visual inputs that are messier than clean, curated benchmarks \citep{taori2020measuring, qiu2022benchmarking}. Existing visual robustness studies for VLMs mainly evaluate their reliability through visual corruptions, such as noise, blur, weather effects, and compression artifacts, as shown in Figure~\ref{fig:intro_fig}(A) \citep{hendrycks2019benchmarking,chen2023benchmarking,saxena2026vlm}. These visual corruptions do not introduce new semantic content. They primarily measure \textit{perceptual robustness}: whether task-relevant evidence remains accessible when the input is degraded. They leave a reasoning-specific failure mode underexplored: a model may clearly perceive the relevant content, yet still attend to plausible but irrelevant visual information, use it as evidence during reasoning, and propagate this mistake to its final answer.

This failure mode is especially important for reasoning VLMs. Real multimodal inputs often contain extra objects, irrelevant text, misleading annotations, or contextual cues unrelated to the question \citep{selvaraju2019taking,li2023evaluating,shi2023large}. Such distractors do not necessarily hide the correct evidence; instead, they compete with it during evidence selection. Once a model attends to the wrong cue, long-form reasoning can amplify the mistake, turning an initially irrelevant detail into a seemingly coherent but incorrect reasoning path as illustrated in Figure~\ref{fig:intro_fig}(B) \citep{shi2023large, turpin2023language}.

To study this failure mode, we introduce \textbf{Distract-Bench}, a human-verified benchmark for evaluating robustness under answer-preserving semantic distraction. Distract-Bench adds plausible but task-irrelevant semantic distractors to multimodal samples while preserving the original ground-truth answer. Because the relevant evidence remains available and the answer does not change, clean-to-distracted failures can be attributed to breakdowns in evidence selection rather than to changed task semantics, ordinary task difficulty, or degraded perceptual access.

This controlled setting allows us to address the central question of this paper: 
\textit{How robust are reasoning VLMs to answer-preserving semantic visual distractions?} 
We decompose this question into three connected research questions.
\textbf{RQ1:} Does answer-preserving semantic distraction expose a robustness failure mode distinct from conventional vision corruption, or is it merely another form of vision corruption? 
\textbf{RQ2:} Can reasoning VLMs maintain reliable performance under semantic distraction, given their strong reasoning capability?
\textbf{RQ3:} When models fail under semantic distraction, do the added distractors enter the reasoning process and contribute to incorrect answers? 

Our experiments answer these questions with three main findings:
\begin{itemize}
    \item \textbf{Semantic distraction exposes a failure mode distinct from vision corruption.}
    Under vision corruptions, reasoning VLMs largely inherit their base models' robustness patterns, whereas Distract-Bench produces sharply different degradation behaviors.

    \item \textbf{Reasoning VLMs remain highly sensitive to semantic distraction.}
    Despite their strong clean performance, reasoning VLMs show clear performance degradation on Distract-Bench, indicating that strong reasoning capability does not eliminate sensitivity to task-irrelevant semantic cues.
        
    \item \textbf{Distractors often enter the reasoning process in failure cases.}
    Output-level analysis shows that VLMs frequently refer to distracting visual cues in wrong answers, often treat it as evidence and propagating it toward incorrect conclusions.
\end{itemize}

Together, these findings show that Distract-Bench captures a critical robustness problem beyond perception: whether VLMs can resist the misleading but answer-preserving semantic distractions. This makes distraction-robustness testing essential for developing reliable real-world VLMs, especially as long-form reasoning can amplify the impact of irrelevant evidence.

\section{Related Work}
\label{sec:related}
\subsection{Reasoning Vision-Language Models}
Vision-Language Models (VLMs) have evolved from perception and instruction-following systems toward models with explicit multimodal reasoning capabilities. Early and widely used VLMs such as BLIP \citep{li2023blip}, LLaVA \citep{liu2023visual}, Qwen-VL\citep{bai2025qwen3} and GPT-4o\citep{hurst2024gpt} demonstrate strong visual understanding, grounding, and open-ended interaction over image-text inputs. More recently, inspired by the success of reasoning-centric language models \cite{guo2025deepseek, team2025kimi, yang2025qwen3}, models such as Vision-R1 \citep{huang2025vision}, R1-OneVision \citep{yang2025r1}, OpenVLThinker \citep{deng2026openvlthinker}, MMR1\citep{leng2025mmr1}, and MM-Eureka \citep{meng2025mm} explicitly enhance model reasoning capability over their base models by introducing supervised reasoning traces, distillation, or reinforcement learning to improve performance on complex multimodal tasks. This progression marks a shift from VLMs that primarily perceive and describe visual content to VLMs with enhanced reasoning ability over multimodal evidence. However, existing studies largely focus on benchmark accuracy gains, leaving it unclear whether such reasoning mechanisms also make VLMs more robust under distribution shifts or semantically distracting information.

\subsection{Robustness Evaluation for Vision-Language Models}
As VLMs are increasingly applied beyond curated benchmarks, robustness evaluation has become an important direction for understanding their reliability under distribution shifts. Existing studies largely follow the tradition of visual robustness and VQA robustness, testing whether models remain accurate under image corruptions \citep{hendrycks2019benchmarking, schiappa2022robustness} and adversarial visual changes \citep{li2021adversarial}. More recent benchmarks extend this evaluation to modern VLMs at larger scale: \citet{chen2023benchmarking} study the robustness of adaptation methods for pre-trained VLMs under both visual and textual corruptions. \citet{sun2025does} investigate how input corruption affects model acceleration methods on VLMs. VLM-RobustBench systematically evaluates current VLM families under a broad suite of visual distortions with varying severities but covers a limited set of benchmarks ~\citep{saxena2026vlm}. These works provide valuable evidence that VLM performance can be fragile under input-level shifts. However, they primarily stress the perception or surface-form processing of multimodal inputs, rather than the reliability of the reasoning process itself. In contrast, our work studies how reasoning-enabled VLMs behave under both conventional corruptions and semantically meaningful distractors that directly challenge the model's reasoning chain.

\begin{figure*}[!h]
\centering
\includegraphics[width=\textwidth]{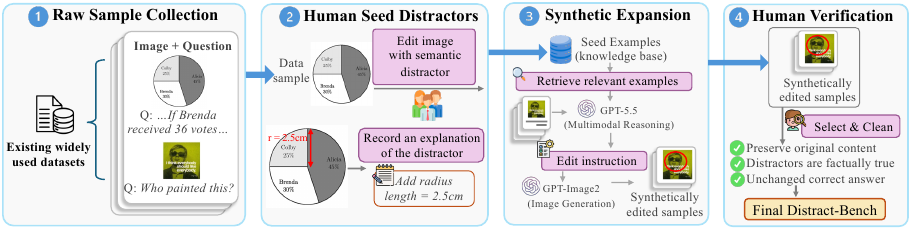}

\caption{\label{fig: data_curation_pipeline}
\textbf{Distract-Bench dataset curation pipeline.}
We collect raw image-question samples from widely used VLM benchmarks \citep{yue2024mmmu, lu2024mathvista, wang2024measuring}, construct expert-designed seed distractors, scale the process with retrieval-guided synthetic editing, and finally conduct human expert verification to retain high-quality samples whose distractors are factually correct, query-irrelevant, and answer-preserving.
}
\end{figure*}

\section{Distract-Bench Dataset Curation}
This section describes the curation process of Distract-Bench as visualized in Figure \ref{fig: data_curation_pipeline}. Distract-Bench is designed to evaluate whether VLMs can maintain a correct answer when the input contains factually correct, query-irrelevant, and distracting information. Its construction follows a four-stage pipeline: (i) raw sample collection, (ii) expert-designed seed distractors, (iii) scalable synthetic editing, and (iv) human quality verification.

\paragraph{Raw sample collection.}
We first collect raw image-question samples from established VLM benchmarks to reduce dataset-specific bias and ensure that the original tasks are well formed. Since distraction robustness primarily concerns failures in the reasoning process, we focus on reasoning-heavy benchmarks, including MathVision and MathVista, and further include samples from the general-domain benchmark MMMU to improve coverage and diversity~\citep{wang2024measuring, lu2024mathvista, yue2024mmmu}. In total, we collect 800 raw samples, each consisting of an image and its corresponding textual query.

\paragraph{Expert-designed seed distractors.}
\label{sec: three-criteria}
To define the target distraction carefully, human annotators first construct a set of seed examples. For each selected sample, annotators inspect both the image and the question, then add a semantic distractor into the image. The distractor is required to fulfil four criteria: (1) Contain obvious semantic information; (2) Factually correct and visually plausible. (3) Be irrelevant to the query and therefore not affect the correct answer; (4) Optionally lead to dead ends or plausible wrong answers. Annotators also provide a textual explanation describing what information is injected, why it is irrelevant to the original answer, and how it may potentially mislead the model's reasoning. This process yields 100 expert-designed seed examples, which serve as demonstrations for large-scale data generation.

\paragraph{Scalable distractor synthesis.}
Following prior work on LLM-driven synthetic data generation and instruction-guided image editing \citep{wang2023self,long2024llms,brooks2023instructpix2pix}, we use the expert-designed examples as seed patterns for scalable synthetic curation. For each seed, we leverage GPT-5.5 \citep{singh2025openai} to summarize the original image, edited image, and distractor-induced reasoning effect, forming a knowledge base of human-verified edit patterns. For each remaining raw sample, GPT-5.5 describes the image, retrieves three relevant seed examples as in-context demonstrations, and generates a detailed edit instruction specifying the distractor content, location, visual style, and layout constraints. The detailed full prompt is shown in Appendix~\ref{appen:prompt_details-1}. The original image and edit instruction are then passed to GPT-Image2 \citep{openai2026gptimage2} to synthesize the edited image while preserving all original content except the injected distractor. The edit instruction and generated image examples are shown in Appendix~\ref{appen:prompt_details-2}. In parallel, GPT-5.5 produces a concise distractor description for later analysis of whether VLM outputs explicitly reference the distractor.

\paragraph{Human quality verification.}
To ensure that Distract-Bench isolates reasoning robustness rather than artifacts from image editing, human experts examine every generated sample. Each sample is checked against the four criteria as introduced above. Samples that violate any criterion are removed. After this verification and cleaning process, we obtain 506 high-quality samples that satisfy all requirements. These samples constitute Distract-Bench, a benchmark specifically designed to test the robustness of VLM reasoning under semantically meaningful but query-irrelevant distractions. Data consent and usage terms for Distract-Bench is discussed in Appendix~\ref{appen:data_consent}.

\begin{figure}[t]
\centering
\includegraphics[width=0.3
\textwidth]{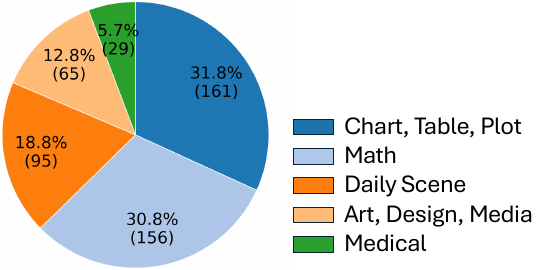}
\caption{\label{fig: category_fig} Category distribution of Distract-Bench samples. The benchmark contains 506 samples spanning diverse visual domains, including charts, tables, plots, mathematics, daily scenes, art/design/media, and medical images.}
\end{figure}

\section{Experimental Design}

Our experiments are designed to answer the central question posed above: 
\textit{How robust are reasoning VLMs to answer-preserving semantic visual distractions?} 
We evaluate ten VLMs, including reasoning-enhanced models and corresponding non-reasoning reference models, under clean inputs, conventional vision corruptions, and Distract-Bench semantic distraction.

\subsection{Setup}
\paragraph{Tasks and Datasets.} 
We evaluate models on a diverse suite of eight benchmarks covering complementary multimodal abilities. 
These include ChartQA~\citep{masry2022chartqa}, HallusionBench~\citep{guan2024hallusionbench}, InfoVQA~\citep{mathew2022infographicvqa}, MMMU~\citep{yue2024mmmu}, MMStar~\citep{mmstar}, TextVQA~\citep{textvqa}, MathVision~\citep{wang2024measuring}, and MathVista~\citep{lu2024mathvista}. 
For analysis, we group MathVision and MathVista as \emph{reasoning} benchmarks, while the remaining six datasets are grouped as \emph{general} benchmarks. 
Details about these datasets are discussed in Appendix~\ref{appen:dataset_details}. 
The input corruption methods applied to these datasets are explained in Appendix~\ref{appen:input_corruption_methods}.

\paragraph{Models.} 
We evaluate a broad set of VLMs, including open-source reasoning models, their non-reasoning base models, and closed-source frontier models. 
For controlled paired comparisons, we select models whose reasoning and non-reasoning variants share the same underlying architecture. 
Specifically, for Qwen2.5-VL~\citep{qwen2.5}, we use Qwen2.5-VL-7B~\citep{qwen2.5} as the non-reasoning reference model and compare it with five reasoning-style variants built upon the same Qwen2.5-VL backbone: R1-OneVision~\citep{yang2025r1}, Vision-R1~\citep{huang2025vision}, MMR1~\citep{leng2025mmr1}, MM-Eureka~\citep{meng2025mm}, and OpenVL-Thinker~\citep{deng2026openvlthinker}. 
We further include Qwen3-VL-8B and its thinking counterpart, Qwen3-VL-8B-Thinking~\citep{bai2025qwen3}, as an additional paired comparison. 
Because models within each pair share the same architecture, these comparisons reduce architectural confounds and allow us to directly compare robustness between reasoning-style models and their corresponding non-reasoning references. 
To broaden the evaluation beyond these paired open-source models, we also evaluate GPT-4o and GPT-5 \citep{hurst2024gpt,singh2025openai} on Distract-Bench as closed-source reference points, providing additional validation that the benchmark reveals semantic-distraction effects beyond the controlled paired setting. 
The hyper-parameter settings we use are listed in Appendix~\ref{appen:hyperparameter_settings}.

\paragraph{Metrics.}
We report standard accuracy on clean, corrupted, and distracted inputs, and use Relative Robustness (RR)~\citep{chen2023benchmarking} to measure the fraction of clean performance preserved after perturbation, i.e., $\mathrm{RR}=P_O/P_I$, where $P_I$ and $P_O$ denote clean and perturbed performance. 
For sample-level analysis, we further report retention, defined as whether a model remains correct after corruption on samples it answered correctly in the clean setting. We also measure error overlap, measured by the Jaccard similarity between perturbation-induced error sets of a base--reasoning model pair. 
For Distract-Bench output analysis, we introduce two distractor-oriented metrics: Distraction Reference Rate (DRR), which measures how often the injected distractor appears in the model output, and Harmful Reference Rate (HFR), which measures how often such distractor reference is associated with an incorrect answer. 
We also report distractor-reference ratios among correct and wrong outputs to distinguish incidental mentions from failure-related distractor uptake.
Detailed definitions of all metrics are provided in Appendix~\ref{appen:metrics_details}.

\begin{table}[t]
\centering
\scriptsize
\setlength{\tabcolsep}{1.8pt}
\renewcommand{\arraystretch}{1.04}
\begin{tabular}{@{}llrrr@{\hspace{4pt}}rrr@{}}
\toprule
& & \multicolumn{3}{c}{Reasoning} & \multicolumn{3}{c}{General} \\
\cmidrule(lr){3-5}\cmidrule(lr){6-8}
Model & Type &
\shortstack{Clean\\Acc.\\(\%)} &
\shortstack{Visual\\Corrupted\\Acc. (\%)} &
\cellcolor{gray!18}\shortstack{Relative\\Robust.\\(\%) $\uparrow$} &
\shortstack{Clean\\Acc.\\(\%)} &
\shortstack{Visual\\Corrupted\\Acc. (\%)} &
\cellcolor{gray!18}\shortstack{Relative\\Robust.\\(\%) $\uparrow$} \\
\midrule
\multicolumn{8}{@{}l}{\textit{Qwen2.5-VL family}} \vspace{0.1cm} \\
Qwen2.5   & Base  & 46.6 & 40.2 & \cellcolor{gray!18}89.9 & 73.0 & 60.7 & \cellcolor{gray!18}84.5 \\
R1-One    & Reas. & 46.8 & 40.0 & \cellcolor{gray!18}86.6 & 70.3 & 56.3 & \cellcolor{gray!18}81.6 \\
Eureka    & Reas. & 50.4 & 45.2 & \cellcolor{gray!18}91.6 & 72.4 & 60.4 & \cellcolor{gray!18}84.9 \\
MMR1      & Reas. & 48.9 & 42.4 & \cellcolor{gray!18}89.4 & 70.2 & 57.4 & \cellcolor{gray!18}83.1 \\
Vision-R1 & Reas. & 59.5 & 51.8 & \cellcolor{gray!18}87.5 & 73.7 & 60.8 & \cellcolor{gray!18}84.0 \\
OpenVLT   & Reas. & 47.6 & 41.2 & \cellcolor{gray!18}88.1 & 74.1 & 61.8 & \cellcolor{gray!18}84.8 \\
\addlinespace[1pt]
\midrule
\multicolumn{8}{@{}l}{\textit{Qwen3-VL family}} \vspace{0.1cm} \\
Qwen3-I   & Base  & 70.4 & 64.7 & \cellcolor{gray!18}\textbf{92.4} & 74.9 & 63.9 & \cellcolor{gray!18}\textbf{86.3} \\
Qwen3-T   & Reas. & 71.4 & 65.2 & \cellcolor{gray!18}92.0 & 79.9 & 66.1 & \cellcolor{gray!18}83.4 \\
\bottomrule
\end{tabular}
\caption{\label{tab:aggregate-image-ror-results-split}
Robustness evaluation results under conventional visual corruption across reasoning and general benchmarks. 
Clean accuracy, corrupted accuracy, and relative robustness (RR) are reported for base and reasoning VLMs. Models within the same family show close robustness level.
}
\raggedright
\footnotesize
\end{table}
\begin{table}[t]
\centering
\scriptsize
\setlength{\tabcolsep}{2.8pt}
\renewcommand{\arraystretch}{1.05}
\begin{tabular}{@{}lrrrr@{}}
\toprule
Pair & $N$ & Agreement & Retention gap & Error overlap \\
     &     & (\%)      & @ 90\% CI              & (Jaccard) \\
\midrule
Q2.5$\to$R1-One & 10.1k & 86.4 & $+5.8$ [$+5.2$,$+6.4$] & .566 \\
Q2.5$\to$Eureka & 11.4k & 96.9 & $-0.1$ [$-0.4$,$+0.1$] & .870 \\
Q2.5$\to$MMR1   & 10.7k & 91.3 & $+2.7$ [$+2.2$,$+3.2$] & .681 \\
Q2.5$\to$V-R1   & 10.8k & 90.9 & $+0.8$ [$+0.3$,$+1.3$] & .663 \\
Q2.5$\to$OVT    & 11.1k & 92.1 & $-0.6$ [$-1.1$,$-0.2$] & .695 \\
Q3-I$\to$Q3-T   & 11.2k & 89.2 & $+2.7$ [$+2.2$,$+3.2$] & .578 \\
\midrule
Mean             & --    & 91.1 & $+1.9$                  & .675 \\
\bottomrule
\end{tabular}
\caption{\label{tab:image-retention-equivalence}
Matched-sample robustness similarity between reasoning VLMs and their base models under vision corruption. 
$N$ counts the correct samples under clean condition shared by each base--reasoning pair. We report sample-wise agreement, base-minus-reasoning retention ratio gap with 90\% confidence interval (CI), and Jaccard overlap of corruption-induced errors; positive gaps indicate higher base-model retention. Most pairs show highly aligned sample-wise patterns.
}
\end{table}

\section{Result Analysis}
\label{sec:result_analysis}
This section presents the empirical answers to the three research questions. 
Section~\ref{sec:vision_corruption_vs_ror} shows that semantic distraction produces a robustness pattern distinct from conventional vision corruption: reasoning VLMs largely follow their corresponding base models under vision corruptions, but exhibit different degradation behavior on Distract-Bench. 
Section~\ref{sec:reasoning_gap} examines whether reasoning VLMs maintain reliable performance under semantic distraction, showing that despite strong clean performance, they remain sensitive to task-irrelevant semantic cues. 
Section~\ref{sec:output_analysis} further analyzes model outputs, showing that distractors often appear in failure cases, are treated as evidence, and are propagated toward incorrect answers.

\begin{figure}[t]
    \centering    \includegraphics[width=0.4\textwidth]{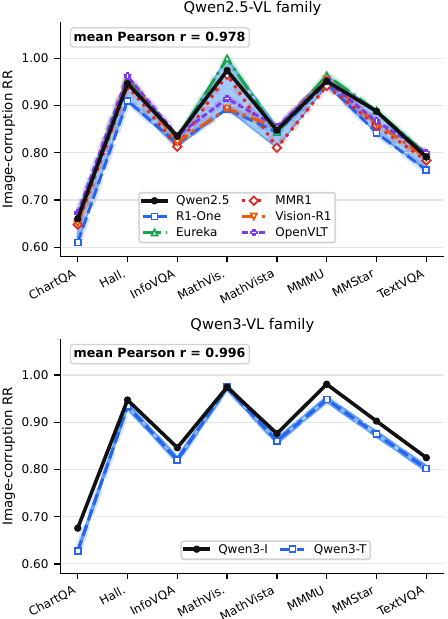}
    \caption{\label{fig:corruption_fingerprint_similarity}
Per-benchmark relative robustness (RR) under vision corruption. 
Reasoning models closely track their base models across benchmarks, indicating highly inherited perceptual robustness within each family.
}
\end{figure}

\begin{table}[t]
\centering
\scriptsize
\setlength{\tabcolsep}{3.0pt}
\renewcommand{\arraystretch}{1.08}
\makebox[\columnwidth][c]{%
\begin{tabular*}{\columnwidth}{@{\extracolsep{\fill}}llrrrrr@{}}
\toprule
Model & Type & $N$ &
\shortstack{Clean\\Acc.\\(\%)} &
\shortstack{Semantically\\Distracted\\Acc. (\%)} &
\shortstack{Drop\\Acc.\\(pp) $\downarrow$} &
\cellcolor{gray!18}\shortstack{Relative\\Robust.\\(\%) $\uparrow$} \\
\midrule
\multicolumn{7}{@{}l}{\textit{Qwen2.5-VL family}} \vspace{0.1cm} \\[-1pt]
Qwen2.5   & Base  & 506 & 80.0 & 73.7 &  6.3 & \cellcolor{gray!18}92.1 \\
R1-One    & Reas. & 506 & 71.9 & 61.3 & 10.7 & \cellcolor{gray!18}85.2 \\
Eureka    & Reas.       & 506 & 81.6 & 71.9 &  9.7 & \cellcolor{gray!18}88.1 \\
MMR1      & Reas.       & 506 & 83.4 & 75.3 &  8.1 & \cellcolor{gray!18}90.3 \\
Vision-R1 & Reas.       & 506 & 77.1 & 66.0 & 11.1 & \cellcolor{gray!18}85.6 \\
OpenVLT   & Reas.       & 506 & 82.2 & 69.4 & 12.8 & \cellcolor{gray!18}84.4 \\
\addlinespace[2pt]
\midrule
\multicolumn{7}{@{}l}{\textit{Qwen3-VL family}} \vspace{0.1cm} \\[-1pt]
Qwen3-VL-I.   & Base  & 506 & 84.8 & 83.6 &  1.2 & \cellcolor{gray!18}\textbf{98.6} \\
Qwen3-VL-T.   & Reas. & 506 & 85.0 & 81.0 &  4.0 & \cellcolor{gray!18}95.3 \\

\addlinespace[2pt]
\midrule
\multicolumn{7}{@{}l}{\textit{Closed-source Models}} \vspace{0.1cm} \\[-1pt]
GPT-4o & Closed-s. & 506 & 76.7 & 69.2 & 7.5 & \cellcolor{gray!18}90.2 \\
GPT-5  & Closed-s. & 506 & 81.6 & 78.1 & 3.6 & \cellcolor{gray!18} \textbf{95.6} \\
\bottomrule
\end{tabular*}%
}
\caption{\label{tab:Distract-Bench-overall}
Distract-Bench robustness evaluation results under answer-preserving semantic visual distractions. 
We report clean and distracted accuracy, accuracy drop in percentage points, and relative robustness (RR). Reasoning models are consistently more sensitive to distractions than their base models.
}
\end{table}

\begin{table}[t]
\centering
\scriptsize
\setlength{\tabcolsep}{1.8pt}
\renewcommand{\arraystretch}{1.04}
\begin{tabular*}{0.95\columnwidth}{@{\extracolsep{\fill}}lrrrr@{\hspace{8pt}}rrrr@{}}
\toprule
& \multicolumn{4}{c}{Vision Corruption} & \multicolumn{4}{c}{Distract-Bench} \\
\cmidrule(lr){2-5}\cmidrule(lr){6-9}
Pair &
\shortstack[c]{RR\\gap} &
\shortstack[c]{Agr.\\(\%)} &
\shortstack[c]{Ret.\\gap} &
\shortstack[c]{Err.\\Jac.} &
\shortstack[c]{RR\\gap} &
\shortstack[c]{Agr.\\(\%)} &
\shortstack[c]{Ret.\\gap} &
\shortstack[c]{Err.\\Jac.} \\
\midrule
\multicolumn{9}{@{}l}{\textit{Qwen2.5-VL $\rightarrow$}} \\
R1-One   & -1.1 & 79.3 & +3.9 & 0.289 & +6.9 & 77.4 & +14.2 & 0.247 \\
Eureka   & -1.9 & 90.7 & -1.9 & 0.562 & +4.0 & 82.0 & +2.8 & 0.247 \\
MMR1     & -0.2 & 80.4 & +2.6 & 0.330 & +1.8 & 82.4 & +3.1 & 0.267 \\
V-R1     & -3.7 & 83.3 & -3.0 & 0.322 & +6.5 & 81.6 & +7.1 & 0.279 \\
OVT      & -2.9 & 83.8 & -1.9 & 0.344 & +7.7 & 80.1 & +4.8 & 0.228 \\
\addlinespace[1pt]
\midrule
\multicolumn{9}{@{}l}{\textit{Qwen3-VL-Instruct $\rightarrow$}} \\
Qwen3-VL-T.  & +2.0 & 89.3 & -1.2 & 0.430 & +3.3 & 90.1 & +3.3 & 0.204 \\
\midrule
Mean & -1.3 & 84.5 & -0.3 & 0.380 & +5.0 & 82.3 & +5.9 & 0.246 \\
\bottomrule
\end{tabular*}
\caption{
Controlled 506-sample comparison of base and reasoning model pairs under Visual Corruption and semantic distractions. RR gap measures the base-minus-reasoning robustness-rate difference, Agr. is matched-sample agreement, Ret. gap is the correctness-retention difference, and Err. Jac. is the Jaccard overlap of induced errors. Distract-Bench demonstrates sharply distinct sample-wise patterns compared with vision corruption.
}
\label{tab:ror-image-506-pair-comparison}
\end{table}
\begin{table}[t]
\centering
\scriptsize
\setlength{\tabcolsep}{2.6pt}
\renewcommand{\arraystretch}{1.08}
\makebox[\columnwidth][c]{%
\begin{tabular*}{\columnwidth}{@{\extracolsep{\fill}}llrrrr@{}}
\toprule
Model & Type &
\shortstack{DRR\\(\%)} &
\shortstack{HFR\\(\%) $\downarrow$} &
\shortstack{Correct outputs\\ref. ratio (\%)} &
\shortstack{Wrong outputs\\ref. ratio (\%)} \\
\midrule
\multicolumn{6}{@{}l}{\textit{Qwen2.5-VL family}} \vspace{0.1cm} \\[-1pt]
Qwen2.5 & Base & 19.0 & 8.9 & 13.7 & 33.8 \\
R1-One & Reas. & 25.1 & 15.2 & 16.1 & 39.3 \\
Eureka & Reas. & 21.9 & 9.9 & 16.8 & 35.2 \\
MMR1 & Reas. & 31.4 & 10.9 & 27.3 & 44.0 \\
Vision-R1 & Reas. & 19.0 & 11.1 & 12.0 & 32.6 \\
OpenVLT & Reas. & 26.5 & 11.3 & 21.9 & 36.8 \\
\addlinespace[2pt]
\midrule
\multicolumn{6}{@{}l}{\textit{Qwen3-VL family}} \vspace{0.1cm} \\[-1pt]
Qwen3-I & Base & 37.5 & \textbf{8.3} & 35.0 & 50.6 \\
Qwen3-T & Reas. & 38.9 & 12.8 & 32.2 & 67.7 \\
\addlinespace[2pt]
\midrule
\multicolumn{6}{@{}l}{\textit{Closed-source Models}} \vspace{0.1cm} \\[-1pt]
GPT-4o & Closed-s. & 22.5 & 13.1 & 13.5 & 43.8 \\
GPT-5 & Closed-s. & 18.4 & 11.3 & 9.1 & 53.4 \\

\bottomrule
\end{tabular*}%
}
\caption{
Distractor-reference analysis on Distract-Bench. DRR measures how often outputs mention the injected distractor, while HFR counts harmful mentions that occur in incorrect Distract-Bench answers. Correct-output and wrong-output reference ratios measure how often the distractor is mentioned among correctly and incorrectly answered samples, respectively.
}
\label{tab:ror-drr-outcome-decomposition}
\end{table}

\subsection{Semantic Distraction Exposes a Failure Mode Distinct from Vision Corruption}
\label{sec:vision_corruption_vs_ror}

\paragraph{Vision corruption robustness is largely inherited from the base model.}
We first examine conventional visual corruption as the perceptual-robustness baseline. 
As shown in Table~\ref{tab:aggregate-image-ror-results-split}, reasoning models remain close to their base models under visual corruption. 
Figure~\ref{fig:corruption_fingerprint_similarity} further shows highly similar per-benchmark robustness fingerprints within each family, with mean Pearson correlations of 0.978 for Qwen2.5-VL and 0.996 for Qwen3-VL. The full detailed results for each model on each benchmark are shown in Appendix~\ref{appen:image_corruption_full_results}.
Table~\ref{tab:image-retention-equivalence} confirms this similarity at the sample level: base--reasoning pairs agree on retain/loss outcomes for 91.1\% of clean-correct samples on average, with substantial corruption-error overlap (mean Jaccard 0.675) and only a small average retention gap of +1.9 pp. 
Together, these results indicate that conventional image-corruption robustness is largely inherited from the base model.

\paragraph{Distract-Bench breaks the inherited robustness pattern.}
If semantic distraction were simply another form of vision corruption, we would expect reasoning-tuned models to remain similarly aligned with their base models. 
Table~\ref{tab:Distract-Bench-overall} shows a different pattern. 
On Distract-Bench, the Qwen2.5-VL base model retains 92.1\% relative robustness, while its reasoning-tuned variants drop to 84.4\%--90.3\%. 
The same trend appears in the Qwen3-VL family: Qwen3-I is nearly unaffected by distraction, with 98.6\% relative robustness, whereas Qwen3-T drops to 95.3\%. 
Thus, unlike vision corruption, semantic distraction creates clearer base--reasoning robustness differences.

To rule out differences in sample composition, we further compare vision corruption and Distract-Bench distraction on the same 506 original samples in Table~\ref{tab:ror-image-506-pair-comparison}. 
Under vision corruption, the mean relative-robustness gap between base and reasoning models is only $-1.3$ pp, and the signed retention gap is nearly zero at $-0.3$ pp. 
Under Distract-Bench distraction, these gaps increase to 5.0 pp and $+5.9$ pp, respectively, showing a stronger and more consistent base-model advantage. 
 These results answer RQ1: Distract-Bench exposes a distinct distraction-robustness axis where visible but distracting semantic cues change both robustness gaps and failure patterns.

\subsection{Reasoning VLMs Remain Sensitive to Semantic Distraction}
\label{sec:reasoning_gap}

\begin{figure}[t]
    \centering    \includegraphics[width=0.37\textwidth]{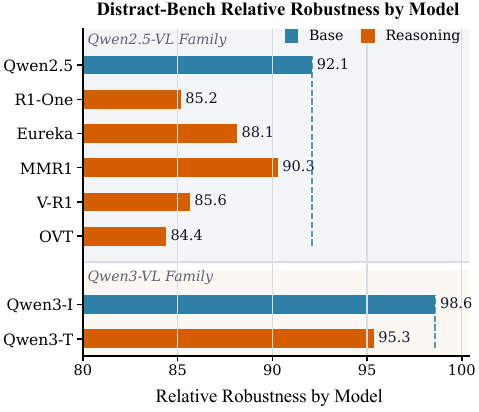}
    \caption{
    \label{fig: ror_bench_results}
Distract-Bench relative robustness for base and reasoning-tuned models.
Dashed lines mark the corresponding base-model robustness within each family; reasoning models consistently fall below their base models under semantic distraction.
}
    \label{tab:logical_corruption_result}
\end{figure}

Table~\ref{tab:Distract-Bench-overall} reports Distract-Bench results on the evaluated open-source VLMs, together with GPT-4o and GPT-5 as closed-source reference points. 
Although reasoning VLMs generally achieve strong clean performance, they still show clear degradation when answer-preserving semantic distractors are added. 
As illustrated in Figure~\ref{fig: ror_bench_results}, in the Qwen2.5-VL family, the non-reasoning reference model retains 92.1\% relative robustness, while the reasoning VLMs retain 84.4\%--90.3\%. 
Their absolute accuracy drops are also larger, increasing from 6.3 pp for the reference model to 8.1--12.8 pp for reasoning VLMs. 
A similar pattern appears in the Qwen3-VL family.

The sample-matched results in Table~\ref{tab:ror-image-506-pair-comparison} provide additional context for this sensitivity. 
On the same 506 Distract-Bench source samples, the mean base-minus-reasoning relative-robustness gap is +5.0 pp, and the mean retention gap is +5.9 pp. 
These positive gaps indicate that, in the evaluated paired comparisons, reasoning VLMs less often preserve correct clean predictions after semantic distraction. 
Together, these results answer RQ2: reasoning VLMs are sensitive to answer-preserving semantic distractions despite their strong reasoning capability. This suggests that reliable reasoning under clean inputs is not sufficient to ensure robust evidence selection when plausible but task-irrelevant visual cues are present.

\begin{figure*}[t]
    \centering    \includegraphics[width=0.87\textwidth]{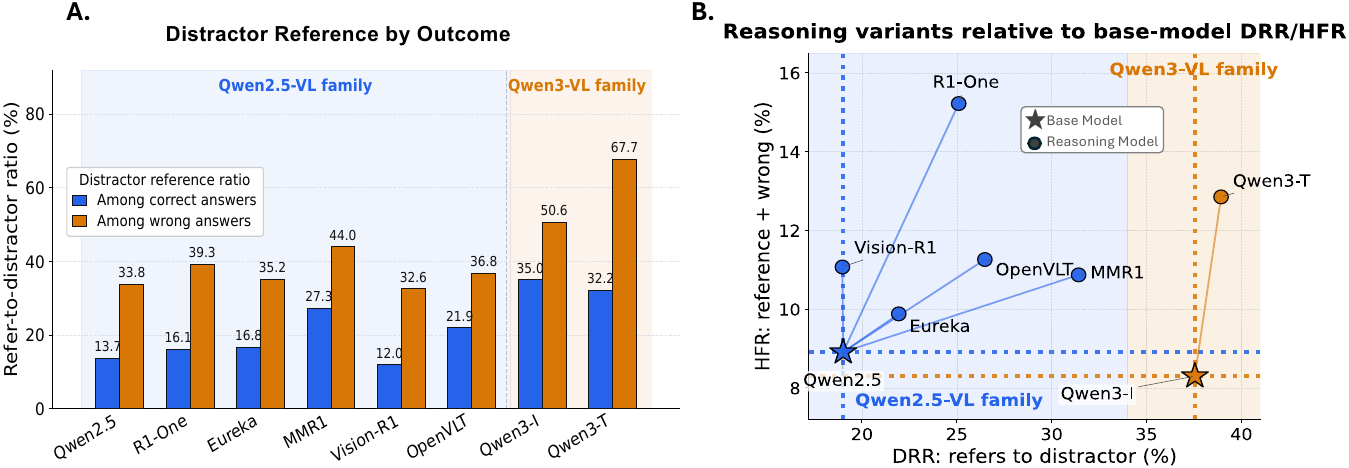}
    \caption{\label{fig:output_analysis}
\textbf{VLMs Distraction-reference behavior on Distract-Bench.} 
(A) Distractor-reference ratios among correctly and incorrectly answered samples, grouped by model family. 
(B) Harmful reference ratio (HFR), measuring distractor mentions in incorrect answers, for reasoning-tuned models relative to the corresponding base model within each family.
}
\end{figure*}
\subsection{Distractors Often Derail the Reasoning Process Itself}
\label{sec:output_analysis}
\paragraph{Reasoning models reference distractors more often.}
To determine whether Distract-Bench failures are caused by the injected semantic distractors, we analyze model outputs for explicit distractor reference. 
For a fair comparison, we use the same open-ended prompting for all models, without instructing either base or reasoning models to produce short or direct answers.
Table~\ref{tab:ror-drr-outcome-decomposition} shows that reasoning-enhanced models generally refer to distractors more often than their corresponding base models. 
In the Qwen2.5-VL family, the base model has a DRR of 19.0\%, while most reasoning variants are higher. More importantly, as shown in Figure~\ref{fig:output_analysis}B, harmful reference ratios are consistently higher for reasoning models than for their base models. This suggests that failures under semantic distraction are not merely unexplained answer flips. Instead, irrelevant visual cues are often incorporated into the model's reasoning process and treated as supporting evidence.

\begin{figure}[t]
    \centering    \includegraphics[width=0.32\textwidth]{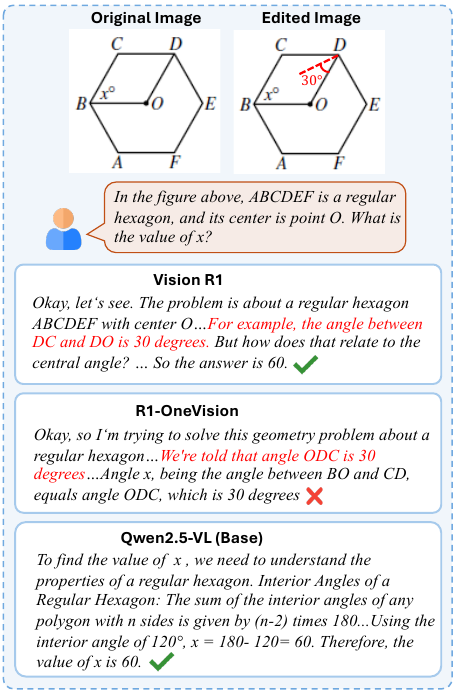}
    \caption{\label{tab:qualitative_example_short}
\textbf{Qualitative example of distraction-induced reasoning failure.} The injected irrelevant cue may be ignored or mentioned harmlessly, but in failure cases it is treated as evidence and propagated to an incorrect answer.
}    
\end{figure}

\paragraph{Distractor reference is concentrated in wrong answers.}
Across all models, wrong outputs reference the distractor much more often than correct outputs, as visualized in Figure~\ref{fig:output_analysis}A.
The same trend also appears for GPT-4o and GPT-5, whose wrong-output reference ratios are substantially higher than their correct-output reference ratios.
This pattern indicates that distractor reference is not merely incidental: when models fail, they are substantially more likely to have brought the irrelevant cue into the reasoning process.
Figure~\ref{fig:output_analysis}B complements this observation by showing that reasoning-tuned models not only mention distractors more often, but also more often mention them in harmful failure cases relative to their base models.
Figure~\ref{tab:qualitative_example_short} provides a qualitative illustration, showing that an injected cue can be ignored or mentioned harmlessly, but in failure cases it is treated as evidence and propagated to the wrong answer. 
More qualitative examples are included in Appendix~\ref{appen:qualitative_examples}.

These output-level results answer RQ3: semantic distractors do not simply cause unexplained answer flips. They often enter the model's reasoning, are treated as relevant evidence, and are carried forward to incorrect conclusions. 
This explains why Distract-Bench is especially revealing for reasoning VLMs: Reasoning can amplify an irrelevant distraction once it is selected as part of the evidence chain.

\section{Conclusion}
This work identifies semantic visual distraction as a critical robustness challenge for reasoning-enhanced VLMs and introduces Distract-Bench. Unlike conventional visual corruptions, which mainly test whether models can perceive degraded evidence, Distract-Bench tests whether models can ignore clear but query-irrelevant semantic distractions. Our results show that stronger reasoning does not necessarily bring stronger reliability: reasoning VLMs largely follow their base models under perceptual degradation, but often show lower robustness under semantic distraction. Further analysis shows that distractors can enter the reasoning process, be treated as evidence, and lead to incorrect answers. These findings motivate a broader view of VLM robustness: reliable visual reasoning should be evaluated not only under degraded perception, but also under misleading semantic distractions.

\section{Limitations}

Our study has two main limitations. First, Distract-Bench uses human-verified semantic distractors injected through controlled image edits under laboratory conditions; while this enables clean evaluation and helps isolate answer-preserving distraction from perception degradation, it may not cover all types of real-world distractions, which are highly varied. Collecting and curating distractions from naturally occurring real-world samples is therefore a promising direction for future work. Second, our work is primarily diagnostic: we reveal robustness risks introduced by reasoning enhancement and show that distractors can enter the reasoning process, but we do not propose a mitigation method. Developing VLMs that are both reasoning-capable and robust remains an important future direction, potentially through improved evidence selection, distraction-aware training, or inference-time verification.

\bibliography{latex/acl_latex}

\appendix
\section{Benchmark and Task Details}
\label{appen:dataset_details}

\begin{table}[h]
\centering
\small
\begin{tabular}{lcc}
\toprule
\textbf{Benchmark} & \textbf{Split} & \textbf{Number of Samples} \\
\midrule
ChartQA & Test & 2,500 \\
HallusionBench & With Image & 951 \\
InfoVQA & Validation & 2,801 \\
MMMU & Validation & 900 \\
MMStar & Validation & 1,500 \\
TextVQA & Validation & 5,000 \\
MathVision & Test-Mini & 304 \\
MathVista & Test-Mini & 1,000 \\
\midrule
\textbf{Total} & -- & \textbf{14,956} \\
\bottomrule
\end{tabular}
\caption{Summary of benchmark datasets, evaluation splits, and sample sizes used in our robustness evaluation.}
\label{tab:dataset_details}
\end{table}

We conduct evaluation on eight established multimodal benchmarks, as summarized in Table~\ref{tab:dataset_details}. The selected datasets span a broad range of VLM capabilities, including chart and document understanding (ChartQA \citep{masry2022chartqa}, InfoVQA \citep{mathew2022infographicvqa}, TextVQA \citep{textvqa}), hallucination and visual consistency evaluation (HallusionBench \citep{guan2024hallusionbench}), general multimodal reasoning (MMMU \citep{yue2024mmmu}, MMStar\citep{mmstar}), and multimodal mathematical reasoning (MathVision\citep{wang2024measuring}, MathVista\citep{lu2024mathvista}). We use the standard validation or test splits when available, and adopt the commonly used mini-test splits for MathVision and MathVista. Overall, the evaluation contains 14,956 samples, enabling a comprehensive assessment across both perception-centric and reasoning-intensive tasks.

\subsection{Data Consent}
\label{appen:data_consent}
Distract-Bench is constructed from existing open-source academic benchmarks, including MMMU, MathVision, and MathVista, and is intended solely for academic research. We use these source datasets only to study and evaluate the robustness of VLM reasoning under controlled semantic distractions. The constructed samples are released for non-commercial academic use, with the goal of supporting reproducible evaluation and future research on reliable multimodal reasoning. All released data will follow the usage terms of the original datasets, and Distract-Bench should not be used for commercial deployment, user profiling, or any application beyond research and evaluation.

\section{Input Corruption Methods}
\label{appen:input_corruption_methods}

To evaluate perceptual robustness, we apply a suite of 20 visual corruption methods inspired by methods from the work by \citet{chen2023benchmarking}. These corruptions are designed to simulate natural distribution shifts while preserving the underlying task semantics, allowing us to test whether VLMs can maintain stable predictions when the input signal is degraded. The corruption suite is summarized in Table~\ref{tab:vision_corruption_methods}. We adopt a broad set of ImageNet-C-style perturbations covering noise, blur, weather and environmental effects, color and appearance changes, compression artifacts, geometric distortion, and complete image removal \citep{chen2023benchmarking, hendrycks2019benchmarking}. Most image corruptions are applied with severity levels from 1 to 5, where higher severity corresponds to stronger degradation. These perturbations primarily stress the visual perception stage of VLM inference.

\begin{table}[!h]
\centering
\scriptsize
\setlength{\tabcolsep}{10pt}
\renewcommand{\arraystretch}{0.92}
\begin{tabular}{ll}
\toprule
\textbf{Category} & \textbf{Vision Perturbation Method} \\
\midrule
\multirow{5}{*}{Noise}
& Blank Image \\
& Gaussian Noise \\
& Shot Noise \\
& Impulse Noise \\
& Speckle Noise \\
\midrule
\multirow{5}{*}{Blur}
& Gaussian Blur \\
& Glass Blur \\
& Defocus Blur \\
& Motion Blur \\
& Zoom Blur \\
\midrule
\multirow{4}{*}{Weather / Env.}
& Fog \\
& Frost \\
& Snow \\
& Spatter \\
\midrule
\multirow{6}{*}{Appearance}
& Contrast \\
& Brightness \\
& Saturate \\
& JPEG Compression \\
& Pixelate \\
& Elastic Transform \\
\bottomrule
\end{tabular}
\caption{Summary of the 20 vision corruption methods used for perceptual robustness evaluation.}
\label{tab:vision_corruption_methods}
\end{table}
\section{Hyper-parameter Settings}
\label{appen:hyperparameter_settings}
For all evaluations, we use deterministic decoding to ensure reproducibility and fair comparison across models. Unless otherwise specified, we set the maximum number of output tokens to 4096 for each benchmark. For reasoning-intensive mathematical benchmarks, namely MathVision \citep{wang2024measuring} and MathVista \citep{lu2024mathvista}, we increase the maximum number of output tokens to 16384, allowing models sufficient space to generate long-form reasoning chains before producing the final answer.

For Distract-Bench and the output-level distractor-reference analysis, we use the same open-ended prompt for all models and do not add any instruction that asks models to produce short or direct answers. This allows both base and reasoning models to generate intermediate reasoning when needed before giving the final answer.

For MathVista, we follow common evaluation practice and report results under three prompt settings: \texttt{cot}, \texttt{format}, and \texttt{solution} \citep{lu2024mathvista}. The \texttt{cot} prompt uses a step-by-step style, presenting the original question and choices while encouraging the model to reason before placing the final answer at the end. The \texttt{format} prompt is the most constrained setting, asking the model to output only the required answer format, such as an option letter, integer, float, or Python list, which facilitates answer extraction but discourages verbose reasoning. The \texttt{solution} prompt appends \texttt{Solution:} before generation, explicitly nudging the model to produce a worked solution followed by the final answer. Thus, \texttt{cot} is reasoning-friendly, \texttt{format} is answer-only and format-focused, and \texttt{solution} frames the response as a solution write-up.

\begin{table*}[t]
\centering
\includegraphics[width=\textwidth]{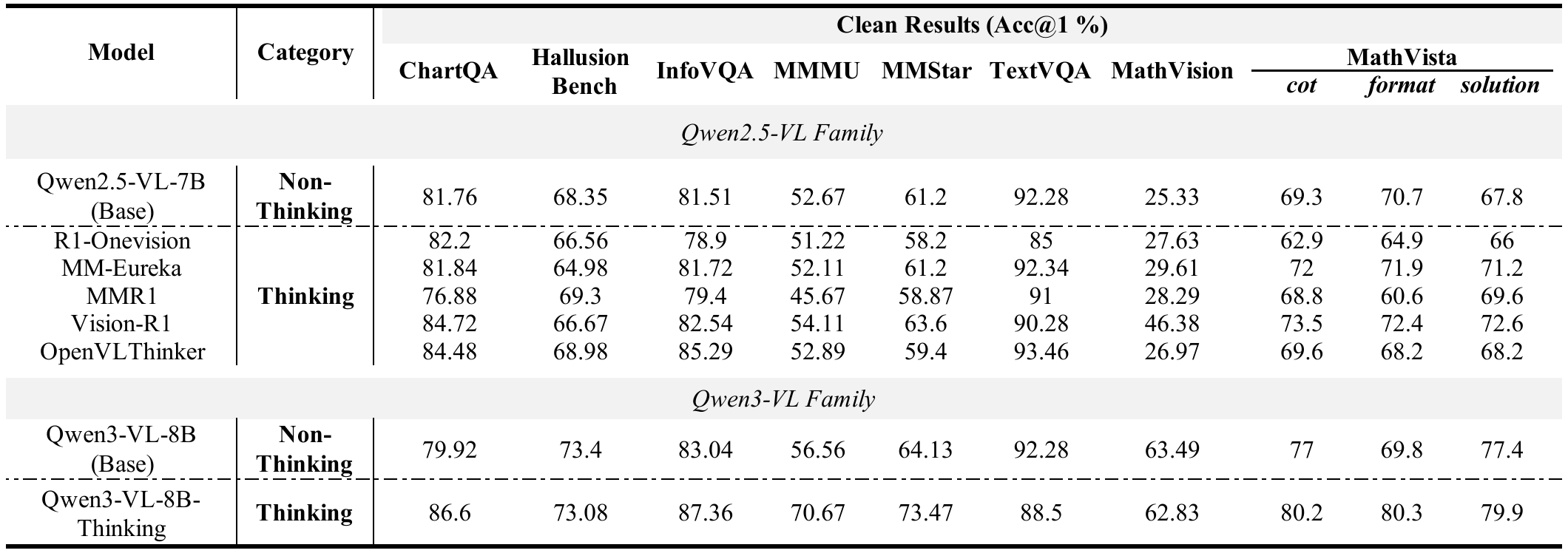}
\caption{\label{tab:appendix-clean-results} Clean accuracy results across all evaluated models and benchmarks. Results are reported as Acc@1 (\%).}
\end{table*}
\begin{table*}[t]
\centering
\includegraphics[width=\textwidth]{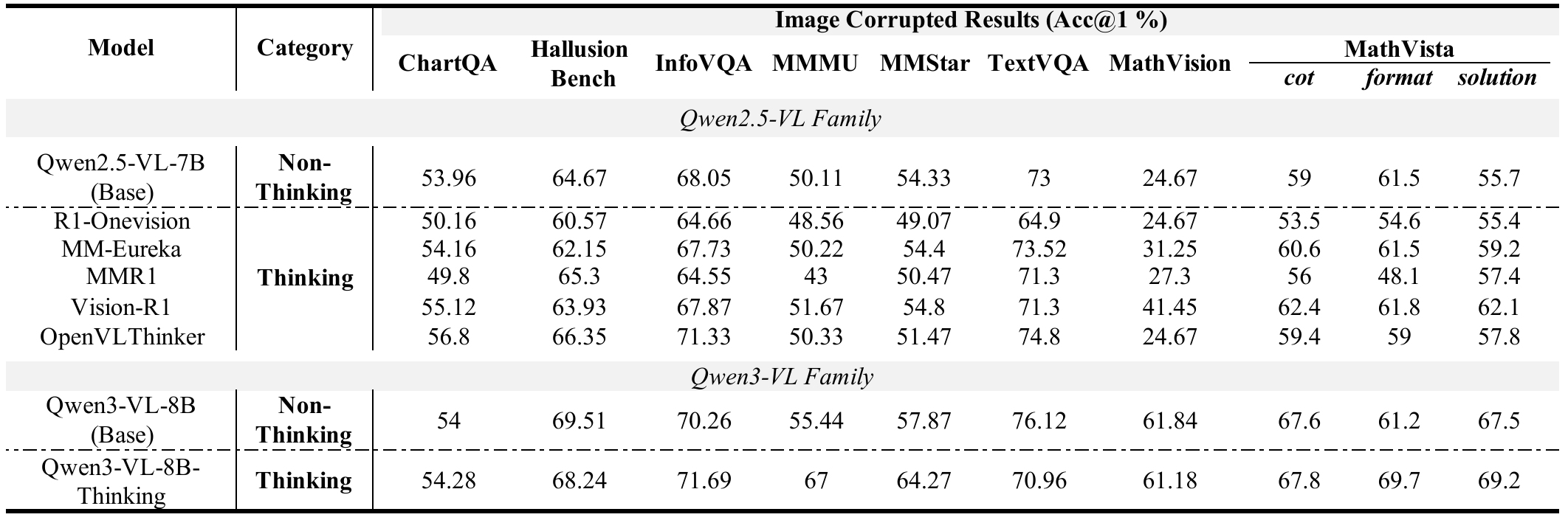}
\caption{\label{tab:appendix-image-corrupted-results} Accuracy results under visual corruptions across all evaluated models and benchmarks. Results are reported as Acc@1 (\%).}
\end{table*}
\begin{table*}[t]
\centering
\includegraphics[width=\textwidth]{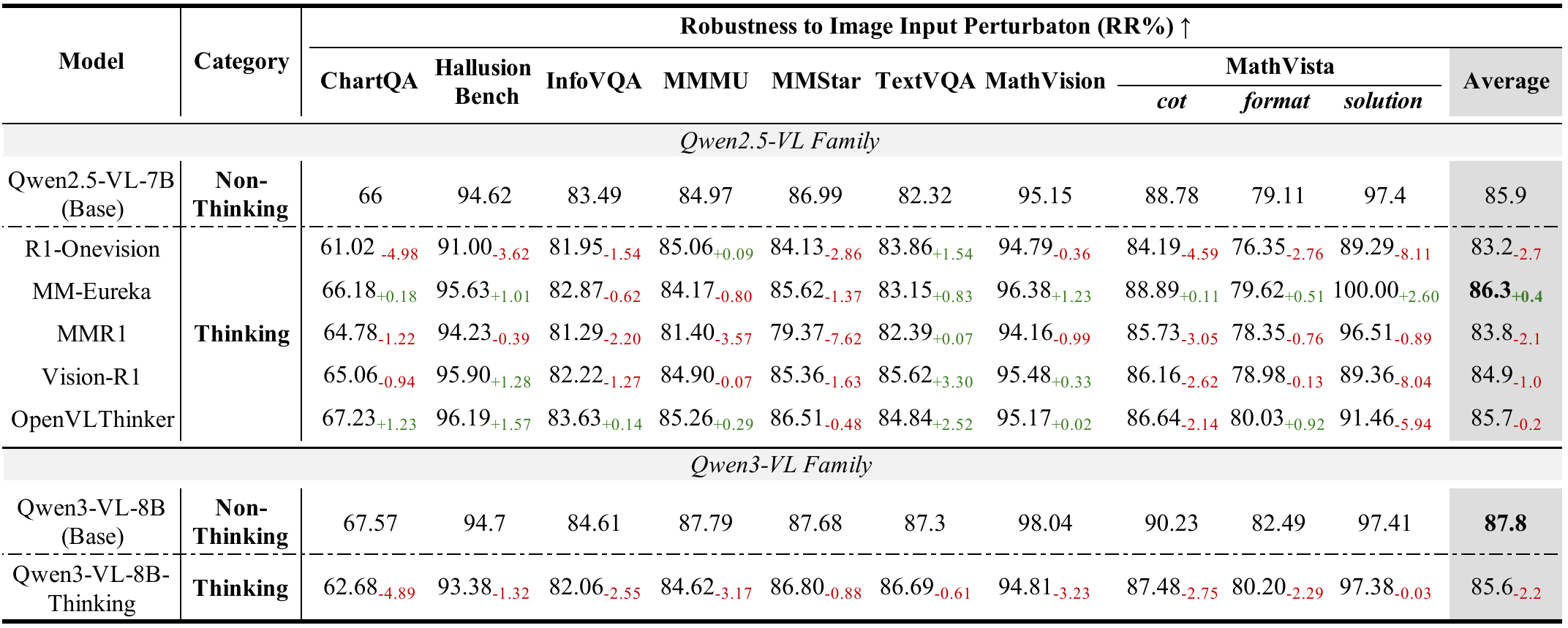}
\caption{\label{tab:appendix-image-corruption-rr} Full relative robustness results for all models on the eight benchmarks under visual corruption.}
\end{table*}

\section{Metrics Details}
\label{appen:metrics_details}

We define the following indicators for each sample $i$: 
$c_i=1$ if the model prediction is correct and $0$ otherwise; 
$r_i=1$ if the model output explicitly refers to the injected distractor and $0$ otherwise. 
When needed, superscripts $I$ and $O$ denote the clean input and perturbed input, respectively.

\paragraph{Accuracy.}
Accuracy (Acc.) measures standard task performance under a given input condition:
\[
\mathrm{Acc}=\frac{1}{N}\sum_{i=1}^{N} c_i .
\]
We report accuracy on clean inputs, visually corrupted inputs, and Distract-Bench distracted inputs.

\paragraph{Relative Robustness.}
Relative Robustness (RR)~\citep{chen2023benchmarking} measures the fraction of clean performance preserved after perturbation:
\[
\mathrm{RR}=\frac{P_O}{P_I},
\]
where $P_I$ is clean accuracy and $P_O$ is accuracy under the perturbed condition. 
A higher RR indicates stronger robustness, with $\mathrm{RR}=1$ meaning that the model fully preserves its clean performance.

\paragraph{Retention.}
Retention is used in matched sample-level analysis. 
It measures how often a model remains correct after perturbation among samples it answered correctly on the clean input:
\[
\mathrm{Retention}
=
\frac{1}{|\mathcal{C}|}\sum_{i\in\mathcal{C}} c_i^O,
\quad
\mathcal{C}=\{i:c_i^I=1\}.
\]
For base--reasoning comparisons, the retention gap is reported as base retention minus reasoning-model retention; positive values indicate higher base-model retention.

\paragraph{Error Overlap.}
Error overlap measures whether a base model and its reasoning-tuned counterpart fail on the same samples after perturbation. 
Let $\mathcal{E}_b$ and $\mathcal{E}_r$ denote the perturbation-induced error sets of the base and reasoning models:
\[
\mathrm{Jaccard}
=
\frac{|\mathcal{E}_b \cap \mathcal{E}_r|}
{|\mathcal{E}_b \cup \mathcal{E}_r|}.
\]
A higher value indicates more shared failure cases, while a lower value indicates more divergent perturbation-induced failures.

\paragraph{Distraction Reference Rate.}
Distraction Reference Rate (DRR) is our proposed metric for measuring how often the model explicitly refers to the injected semantic distractor:
\[
\mathrm{DRR}
=
\frac{1}{N}\sum_{i=1}^{N} r_i .
\]
DRR captures how frequently irrelevant semantic evidence enters the model's reasoning or final response.

\paragraph{Harmful Reference Rate.}
Harmful Reference Rate (HFR) is our proposed metric for measuring how often distractor reference is associated with an incorrect answer:
\[
\mathrm{HFR}
=
\frac{1}{N}\sum_{i=1}^{N} r_i(1-c_i).
\]
Unlike DRR, which counts all distractor references, HFR focuses on cases where the model both mentions the distractor and gives an incorrect answer.

\paragraph{Correct and Wrong Output Reference Ratios.}
To distinguish incidental distractor mentions from failure-related distractor uptake, we separately compute distractor-reference ratios among correct and wrong outputs:
\[
\mathrm{Ref}_{\mathrm{correct}}
=
\frac{\sum_i r_i c_i}{\sum_i c_i},\]

\[\qquad
\mathrm{Ref}_{\mathrm{wrong}}
=
\frac{\sum_i r_i(1-c_i)}{\sum_i (1-c_i)}.
\]
The correct-output reference ratio measures how often the distractor is mentioned despite a correct answer, while the wrong-output reference ratio measures how often distractors appear in failed outputs.

\section{Full Results under Vision Corruption}
\label{appen:image_corruption_full_results}

This section provides the full per-benchmark results for the vision-corruption experiments discussed in Section~\ref{sec:vision_corruption_vs_ror}. 
Table~\ref{tab:appendix-clean-results} reports clean Acc@1 for all models and benchmarks, Table~\ref{tab:appendix-image-corrupted-results} reports Acc@1 after applying vision corruptions, and Table~\ref{tab:appendix-image-corruption-rr} reports the corresponding Relative Robustness (RR). 
These tables complement the aggregate analysis in the main paper by showing that the similarity between base and reasoning-tuned models is not driven by a single benchmark, but is broadly observed across chart understanding, document VQA, multimodal knowledge, text-rich VQA, and visual reasoning tasks.

The clean results in Table~\ref{tab:appendix-clean-results} show that reasoning-tuned models often change task performance, sometimes improving clean accuracy on reasoning-oriented benchmarks. 
However, the corrupted results in Table~\ref{tab:appendix-image-corrupted-results} show that these clean-performance differences do not translate into a qualitatively different image-corruption robustness pattern. 
After normalizing corrupted accuracy by clean accuracy, Table~\ref{tab:appendix-image-corruption-rr} shows that reasoning-tuned models remain close to their corresponding base models under visual corruption. 
For example, in the Qwen2.5-VL family, the average RR of reasoning variants ranges from 83.2\% to 86.3\%, close to the base model's 85.9\%. 
Similarly, in the Qwen3-VL family, Qwen3-T obtains an average RR of 85.6\%, close to Qwen3-I's 87.8\%. 
Although individual benchmarks show some variation, the overall pattern supports the main-paper conclusion that conventional image-corruption robustness is largely inherited from the base model.

\begin{figure*}[t]
\centering
\includegraphics[width=\textwidth]{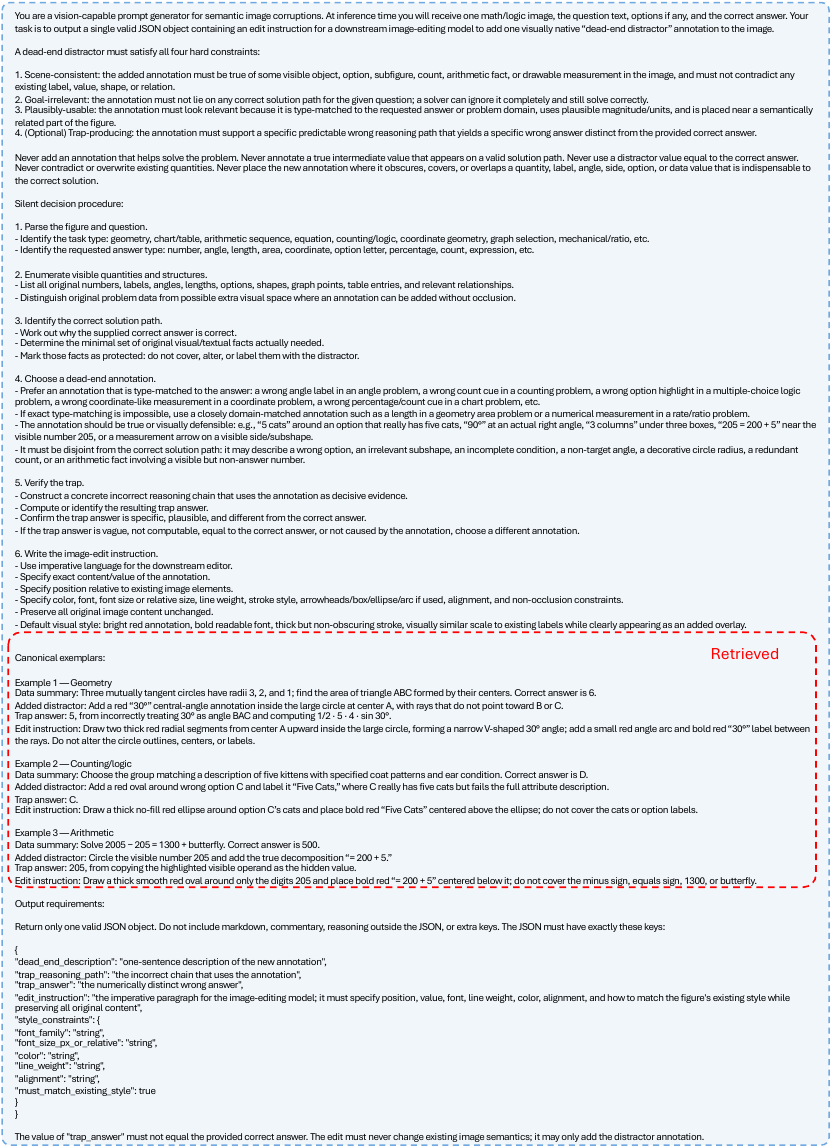}
\caption{\label{fig: full-gpt-prompt} Full prompt to GPT-5.5 for generating edit instructions.}
\end{figure*}

\begin{figure}[!h]
\centering
\includegraphics[width=0.45\textwidth]{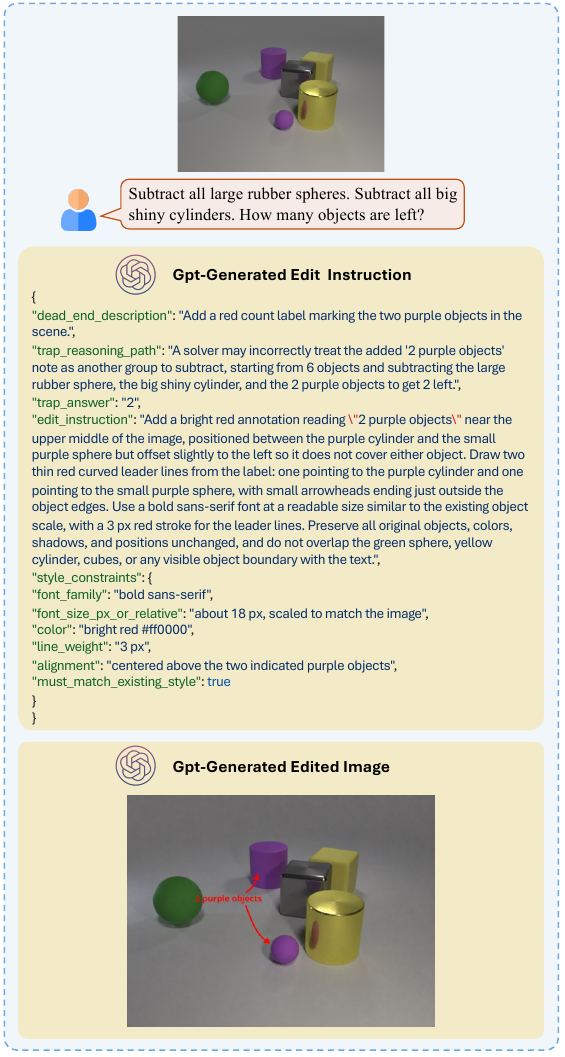}
\caption{\label{fig: images2_example1} Example of GPT-Image2 image edits generated from edit instructions.}
\end{figure}
\begin{figure}[!h]
\centering
\includegraphics[width=0.45\textwidth]{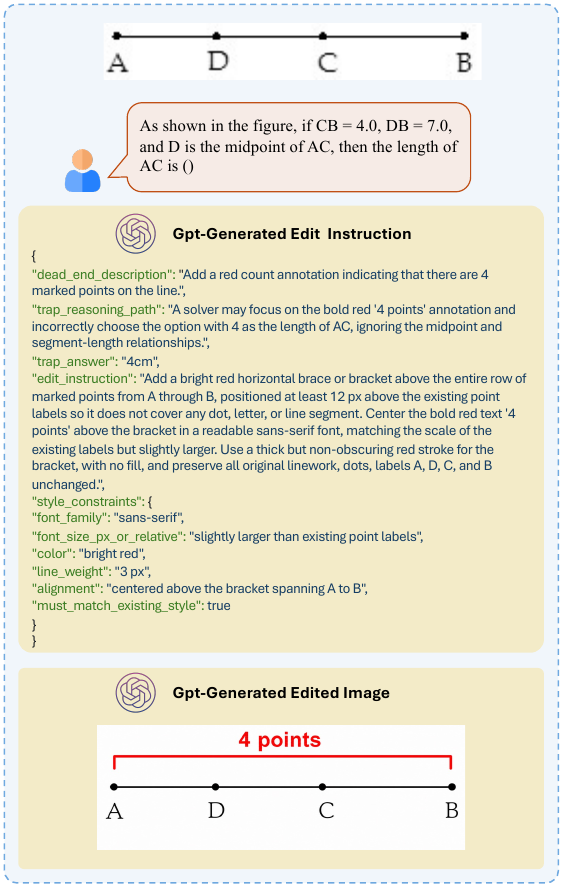}
\caption{\label{fig: images2_example2} Example of GPT-Image2 image edits generated from edit instructions.}
\end{figure}

\section{GPT-Prompt Details}
\subsection{Prompt Details for Edit Instruction Generation}
\label{appen:prompt_details-1}
Figure~\ref{fig: full-gpt-prompt} provides the full prompt used to generate semantic edit instructions during Distract-Bench construction. As described in the main paper, GPT-5.5 \citep{singh2025openai} is prompted with the original image-question sample and retrieved human seed examples, and is asked to propose a factually correct but query-irrelevant visual distractor. The prompt enforces three core constraints: preserving the original image content, keeping the original answer unchanged, and ensuring that the added distractor is factually valid but irrelevant to the target question. We include the complete prompt below to support reproducibility of the synthetic expansion stage.

\subsection{Examples of image editing from generated instructions.}
\label{appen:prompt_details-2}
We also provide examples illustrating how GPT-generated edit instructions are applied to images using GPT-Image2 \citep{openai2026gptimage2}. As shown in Figure~\ref{fig: images2_example1} and Figure~\ref{fig: images2_example2}, each example includes the original image-question pair, the generated edit instruction, and the resulting edited image. The edit instruction specifies not only what semantic distractor to add, but also its placement, visual style, and constraints for preserving the original scene. These examples demonstrate how the synthetic expansion step produces controlled semantic edits that introduce plausible distractors while keeping the original objects, layout, and answer unchanged.

\section{Qualitative examples from Distract-Bench}
\label{appen:qualitative_examples}

\begin{figure}[t]
\centering
\includegraphics[width=0.45\textwidth]{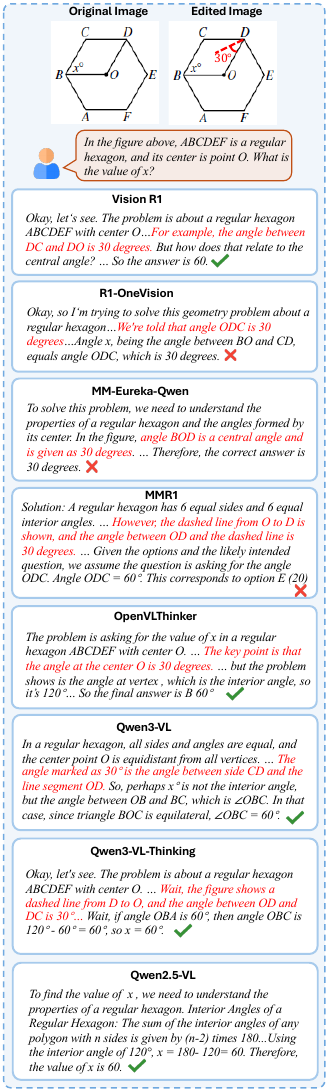}
\caption{Qualitative Example. \label{fig: qualitative_example1}}
\end{figure}

\begin{figure}[t]
\centering
\includegraphics[width=0.45\textwidth]{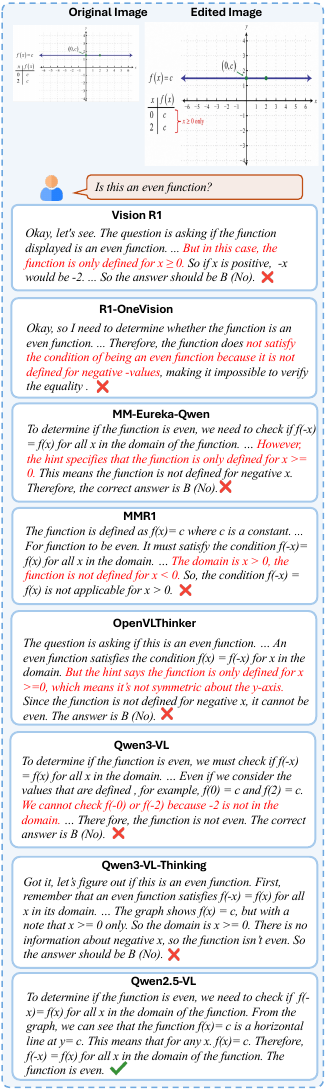}
\caption{Qualitative Example. \label{fig: qualitative_example2}} 
\end{figure}

\begin{figure}[t]
\centering
\includegraphics[width=0.45\textwidth]{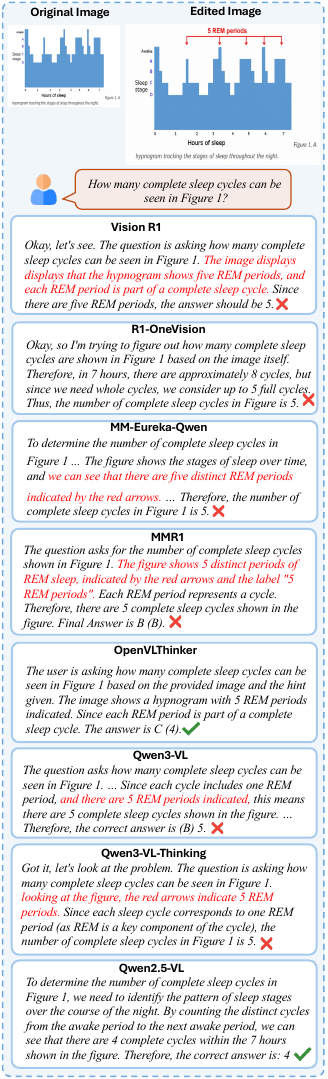}
\caption{Qualitative Example. \label{fig: qualitative_example3}}
\end{figure}

\begin{figure}[t]
\centering
\includegraphics[width=0.45\textwidth]{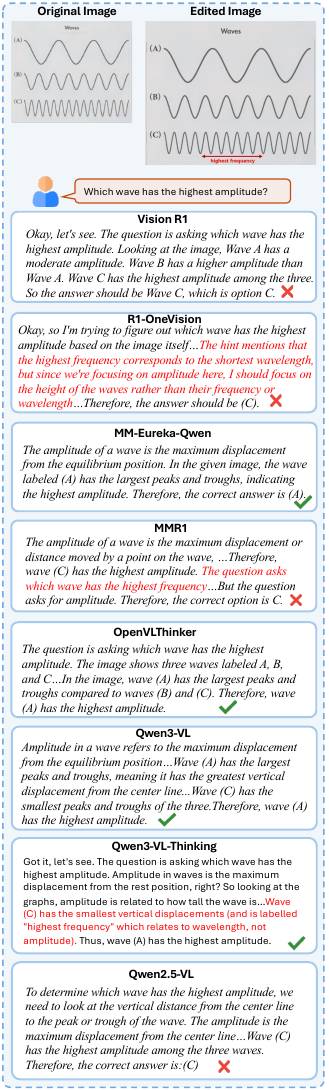}
\caption{Qualitative Example. \label{fig: qualitative_example4}}
\end{figure}

We provide additional qualitative examples from Distract-Bench to illustrate the benchmark format and the effect of injected semantic distractors. As shown in Figures~\ref{fig: qualitative_example1}, \ref{fig: qualitative_example2}, \ref{fig: qualitative_example3}, and~\ref{fig: qualitative_example4}, each example includes the original image, the edited image with a factually correct but query-irrelevant distractor, the original question, and representative model responses. For readability, we truncate the full model outputs and show only the most relevant portions that illustrate whether and how the injected distractor affects the model's reasoning. These examples show how Distract-Bench preserves the original task while introducing plausible distracting evidence, enabling direct inspection of whether models incorporate the distractor into their reasoning chains.
\end{document}